\documentclass[conference]{IEEEtran}
\IEEEoverridecommandlockouts
\UseRawInputEncoding
\usepackage[utf8]{inputenc}
\usepackage{cite}
\usepackage[labelfont=bf]{caption}
\usepackage{amsmath,amssymb,amsfonts}
\usepackage{algorithmic}
\usepackage{graphicx}
\usepackage{textcomp}
\usepackage{xcolor}
\usepackage{booktabs}
\usepackage{url}
\usepackage{float}
\usepackage{comment}
\def\BibTeX{{\rm B\kern-.05em{\sc i\kern-.025em b}\kern-.08em
    T\kern-.1667em\lower.7ex\hbox{E}\kern-.125emX}}
\begin{document}

\title{CT-VDETR: Semi-supervised 3D Trauma Detection in Computed Tomography (CT) scans using Dense Vertex Relative Position Encoding\\
}

\author{
    \IEEEauthorblockN{Shivam Chaudhary}
    \IEEEauthorblockA{\textit{Pattern Recognition Lab}\\
    \textit{FAU Erlangen-Nürnberg}\\
    shivam.chaudhary@fau.de}
    \and
    \IEEEauthorblockN{Sheethal Bhat}
    \IEEEauthorblockA{\textit{Pattern Recognition Lab}\\
    \textit{FAU Erlangen-Nürnberg}\\
    sheethal.bhat@fau.de}
    \and
    \IEEEauthorblockN{Andreas Maier}
    \IEEEauthorblockA{\textit{Pattern Recognition Lab}\\
    \textit{FAU Erlangen-Nürnberg}\\
    andreas.maier@fau.de}
}

\maketitle
\begin{abstract}

Accurate detection and localization of traumatic injuries in abdominal CT remain challenging because voxel-level annotations are limited and expensive to obtain. We present a label-efficient framework for 3D abdominal trauma detection that combines self-supervised pretraining with semi-supervised transformer-based detection. First, we use Masked Image Modeling (MIM) on 1,098 CT volumes to pretrain a 3D U-Net encoder for anatomical representation learning. Next, we adapt V-DETR to dense volumetric CT through a feature adapter that converts the encoder feature grid into a compact token sequence for transformer decoding. The pretrained encoder is then integrated with V-DETR and 3D Vertex Relative Position Encoding (3DV-RPE) to improve the localization of irregularly shaped injuries. Finally, semi-supervised teacher--student consistency regularization leverages 2,000 additional unlabeled volumes during detector training. To the best of our knowledge, this is the first application of a 3D DETR-style detector to the RSNA abdominal trauma detection task. On this benchmark, the proposed method achieves 31.33\% test mAP@0.50 using only 78 labeled training volumes, corresponding to a 1.53$\times$ improvement over supervised-only training. These results show that combining medical-domain pretraining with semi-supervised learning is an effective strategy for label-scarce 3D medical detection.
\end{abstract}

\begin{IEEEkeywords}
3D Object Detection, Self-supervised learning, Semi-Supervised Learning, Transformer, DETR, Relative Position Encoding, U-Net, Computed Tomography
\end{IEEEkeywords}

\section{Introduction}
Medical imaging plays a pivotal role in modern diagnostic workflows, with Computed Tomography (CT) scans being particularly valuable for detecting internal injuries and anatomical abnormalities, especially in trauma and emergency care \cite{brenner2007ct}. Manual analysis of 3D medical volumes is time-intensive, requires expert knowledge, and is prone to inter-observer variability \cite{elmore2009variability}. The development of automated 3D object detection systems for medical imaging promises to improve diagnostic accuracy, reduce radiologist workload, and enable faster clinical decision-making, particularly in emergency settings where timely diagnosis is critical \cite{litjens2017survey}. 

Prior approaches to medical image analysis have predominantly relied on 2D slice-wise evaluation, patch or sliding window-based strategies, or computationally expensive 3D convolution networks \cite{shen2017deep}. In parallel, progress in supervised 3D learning is limited by the scarcity of large, publicly available, expertly annotated medical datasets. This limitation is particularly relevant for voxel or pixel-level annotations such as segmentation masks, whose creation requires substantial expert effort \cite{tajbakhsh2020embracing}. 

To address this limitation, self-supervised learning (SSL) has emerged as an effective strategy for exploiting large amounts of unlabeled medical data to learn transferable features \cite{taleb20203d}. Among SSL approaches, masked image modeling (MIM) and masked autoencoding have shown strong performance by training models to reconstruct masked portions of the input, thereby encouraging the learning of semantically meaningful structure. The Masked Autoencoder (MAE) framework of He et al.~\cite{he2022mae} demonstrated the effectiveness of this strategy in natural images, and subsequent work has explored analogous pretraining paradigms for volumetric medical data \cite{zhou2021models}. In the context of 3D medical detection, Eckstein et al.~\cite{eckstein2024missing} showed that self-supervised pretraining on unlabeled medical volumes can substantially improve downstream detection performance, particularly in low-label regimes.

Complementary to SSL, semi-supervised learning aims to exploit unlabeled samples during task-specific training by combining a small labeled set with a larger unlabeled set. Teacher--student frameworks are a common strategy in this setting. Mean Teacher~\cite{tarvainen2017mean}, for example, uses an exponential moving average (EMA) model as a teacher and enforces prediction consistency under perturbations. For object detection, Unbiased Teacher~\cite{liu2021unbiased} extended this paradigm to mitigate confirmation bias in pseudo-labeling and demonstrated strong results in natural-image detection. Related semi-supervised approaches have also been explored in medical imaging, where annotation scarcity makes such strategies especially attractive \cite{cheplygina2019notso}. However, the combination of self-supervised pretraining, semi-supervised optimization, and transformer-based 3D detection remains relatively underexplored in CT.

Anatomical structures in medical CT exhibit significant variability in shape, size, and location across patients due to individual demographics, pathology, and acquisition protocols, unlike the relatively rigid objects common in natural image datasets. This inherent heterogeneity poses a fundamental challenge for developing robust and generalizable 3D detection systems in medical imaging.

Developing robust 3D detection systems for medical imaging is further complicated by heterogeneity in volumetric acquisition, including variations in slice thickness, voxel spacing, field of view, and reconstruction kernels, as well as the need to model anatomical structures over their full 3D extent \cite{litjens2017survey}. Moreover, full-volume 3D processing is computationally demanding and often exceeds practical memory and runtime limits, especially at high spatial resolution \cite{cciccek20163dunet}. These factors make it difficult to design methods that are both computationally feasible and sufficiently expressive for volumetric reasoning.

In parallel, 3D object detection has evolved from convolutional extensions of 2D detectors toward set-prediction and transformer-based formulations. Early volumetric adaptations of detectors such as Faster R-CNN~\cite{ren2015faster} replaced 2D operators with 3D convolutions, but these methods generally remain computationally expensive and often require careful design of anchors and other task-specific hyperparameters \cite{zhou2018voxelnet}. More recent 3D detection methods such as VoteNet~\cite{qi2019votenet}, 3DETR~\cite{misra20213detr}, and GroupFree3D~\cite{liu2021groupfree} were developed primarily for point-cloud scene understanding. DETR~\cite{carion2020detr} in particular established a set-prediction view of detection, replacing heuristic components such as anchor generation and non-maximum suppression with object queries and bipartite matching. While these developments are conceptually appealing, direct transfer to CT remains non-trivial because medical volumes are represented as dense voxel grids rather than native point clouds, and because existing geometric encodings are not designed for modality-specific anatomical structure \cite{shen2023vdetr}. 

A relevant recent direction is V-DETR~\cite{shen2023vdetr}, which improves geometric reasoning in 3D detection by introducing vertex-relative positional encoding (3DV-RPE) instead of relying only on center-based box representations. This design provides richer spatial cues for object localization and achieves strong performance on large-scale indoor point-cloud benchmarks such as ScanNetV2~\cite{dai2017scannet}. However, its target setting differs substantially from 3D CT, where data are volumetric, annotations are scarce, and learned representations must capture medical image appearance rather than sparse scene geometry. 
More broadly, methods adapted directly from 2D detection are often not optimal for 3D CT, since compact 2D box representations may not adequately capture the extent of elongated, deformable, or irregular anatomical structures and lesions across slices \cite{eckstein2024missing}. Finally, transfer learning from natural-image or video pretraining is often less effective in medical imaging because the data distribution differs substantially, including grayscale appearance, modality-specific intensity characteristics such as Hounsfield units in CT, and highly specialized anatomical patterns \cite{tajbakhsh2020embracing}. These differences motivate the development of architectures and representations that are tailored specifically to volumetric medical data.

\begin{figure*}[htbp]
\centering
\includegraphics[width=\textwidth]{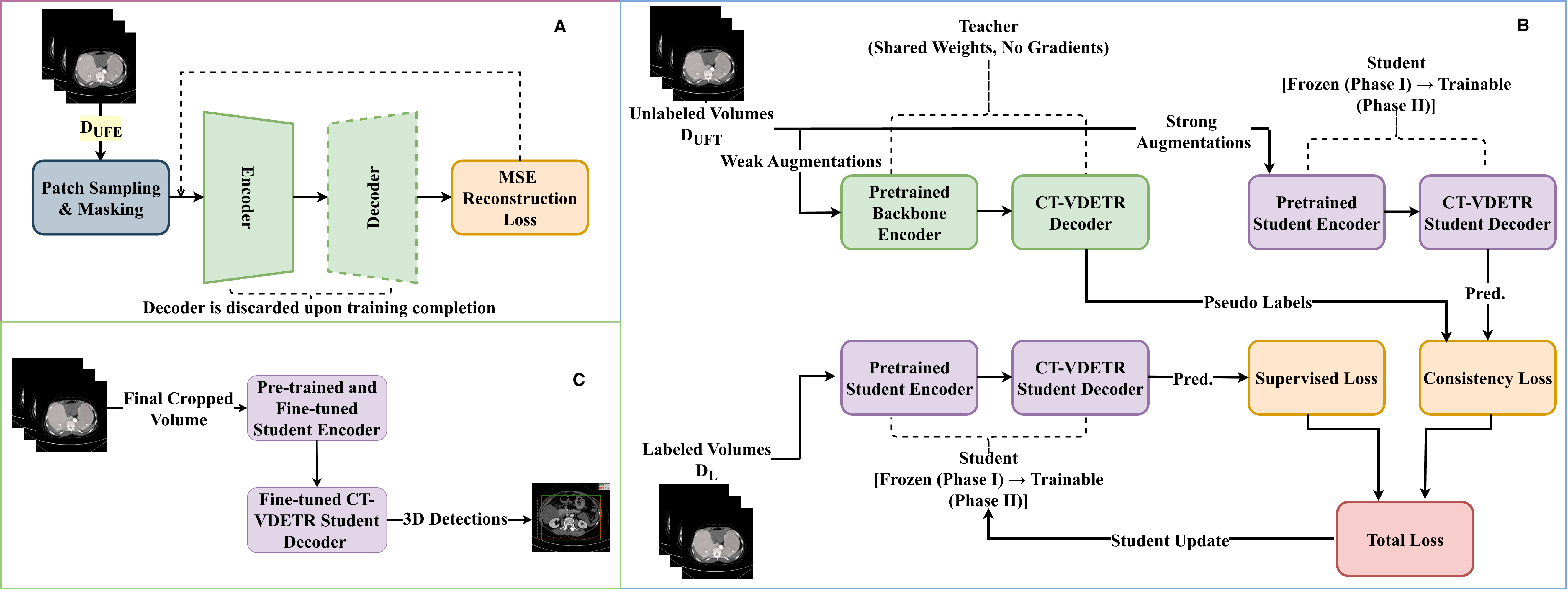}
\caption{\textbf{Overview of the proposed method for 3D CT-VDETR.} (A) Self-supervised pre-training of 3D U-Net-based feature extractor backbone. (B) Semi-supervised fine-tuning of the CT-VDETR framework. (C) Inference stage.}
\label{fig:overview}
\end{figure*}

In this paper, we propose a novel transformer-based 3D detection framework for abdominal trauma CT on the RSNA 2023 Abdominal Trauma Detection dataset \cite{rsna2023_abdominal_trauma}. Fig.~\ref{fig:overview} summarizes the pre-training and fine-tuning pipeline. In the pre-training stage, we use a 3D U-Net as a feature extractor and train it in a self-supervised manner using a reconstruction MSE loss. After pre-training, the decoder is discarded, and the encoder is used as the feature extraction backbone. This pre-trained backbone is then integrated into the proposed CT-VDETR framework and fine-tuned using a semi-supervised strategy. The student model is trained with supervised detection loss on labeled volumes and a consistency loss on unlabeled data, where a teacher branch evaluated on a weakly augmented view provides pseudo-labels. To stabilize training, a two-phase strategy is adopted: in Phase I, the encoder is frozen, and only the detector is trained; in Phase II, the encoder is unfrozen, and the entire network is fine-tuned, using a lower learning rate for the encoder. During inference, only the trained student model is used to predict 3D bounding boxes and class scores.

\subsection{Contributions:} 

This study makes three main contributions. 1) We employ Masked Image Modeling to pre-train a 3D U-Net encoder on 1,098 unlabeled CT volumes, enabling anatomical representation learning without manual annotations. 2) We introduce a novel feature adapter that enables V-DETR's sparse point cloud architecture to process dense 3D CT volumes by sampling representative tokens from the encoder's feature grid, thereby bridging the gap between natural 3D detection and medical volumetric imaging. 3) Finally, we leverage 2,000 unlabeled volumes through teacher--student consistency regularization during detection training, achieving a 1.53$\times$ improvement on the RSNA test dataset over supervised-only training while preventing performance collapse in low-label settings.

\section{Dataset details}

\subsection{Dataset Overview}

The RSNA Abdominal Trauma Detection dataset~\cite{rsna2023_abdominal_trauma} comprises 4,711 CT series stored in DICOM format with substantial variation in the number of slices per series due to differences in acquisition protocol and anatomical coverage. In our setup, we distinguish between series with voxel-level injury annotations and those without such annotations. Specifically, 206 series (4.4\%) provide abdominal injury segmentation masks in NIfTI format, while the remaining 4,505 series (95.6\%) do not include voxel-level injury masks and are treated as unlabeled data in our framework. To clarify, the unlabeled scans correspond to cases without annotated abdominal injury.

From these unlabeled, non-injury scans, we define two subsets. $\mathcal{D}_{UFE}$ denotes the subset used for self-supervised pretraining of the feature extractor, $\mathcal{D}_{UFT}$ denotes the subset used as unlabeled data during semi-supervised detector training with samples not included in $\mathcal{D}_{UFE}$. From the 206 scans with injury annotations, as a preliminary study, we extract liver-specific injuries only, yielding 123 scans that form $\mathcal{D}_{L}$, representing the subset with voxel-wise liver injury annotations employed for supervised detection training. For the detection task, the provided liver segmentation masks are converted into 3D bounding-box targets. Table~\ref{tab:dataset_stats} summarizes the composition of these subsets.

This partitioning supports the proposed two-stage training strategy: self-supervised pretraining of the encoder using $\mathcal{D}_{UFE}$, followed by semi-supervised detector training using $\mathcal{D}_{L}$ together with unlabeled volumes from $\mathcal{D}_{UFT}$. $\mathcal{D}_{L}$ is randomly split into 78 training, 20 validation, and 25 test volumes.

\begin{table}[htbp]
\caption{Preprocessed Dataset Statistics}
\label{tab:dataset_stats}
\centering
\begin{tabular}{lcc}
\hline
\textbf{Category} & \textbf{Count} & \textbf{Volume Size} \\
\hline
Unlabeled Vol. Dataset ($\mathcal{D}_{UFE}$)  & 1,000 & $512 \times 336 \times 336$ \\
Unlabeled Vol. Dataset ($\mathcal{D}_{UFT}$) & 2,000 & $512 \times 336 \times 336$ \\
Labeled Vol. Dataset ($\mathcal{D}_{L}$) & 123 & $512 \times 336 \times 336$  \\
Total Series & 3,123 & --  \\
\hline
Voxel Spacing $(Z, Y, X)$ & -- & 2.0, 1.0, 1.0 mm  \\
Intensity Range & -- & $[0, 1]$  \\
HU Window & -- & $[-100, 300]$  \\
\hline
\end{tabular}
\end{table}

\subsection{Preprocessing Pipeline for Unlabeled Volumes}

DICOM slices are loaded, sorted along the z-axis, and stacked into 3D arrays with dimensions (Z, Y, X). Raw pixel intensities are converted to Hounsfield Units (HU) using DICOM rescale parameters:

\begin{align}
HU = I_{raw} \cdot m + b,
\end{align}
where $I_{raw}$ is the raw pixel value, $m$ is RescaleSlope, and $b$ is RescaleIntercept. Intensities are clipped to [-100, 300] HU (abdominal soft tissue window) and normalized to [0, 1] via min-max normalization:

\begin{align}
I_{norm} = \frac{I - I_{min}}{I_{max} - I_{min}},
\end{align}
with $I_{min} = -100$ HU and $I_{max} = 300$ HU. 


Next, volumes are resampled to a voxel spacing of $(2.0, 1.0, 1.0)$ mm using trilinear interpolation and standardized to a fixed size of $512 \times 336 \times 336$ voxels through cropping and zero-padding. This target size was chosen to cover the majority of volumes while limiting unnecessary truncation. Preprocessed volumes are stored in compressed \texttt{.npz} format. The same preprocessing pipeline is applied to the unlabeled subsets $\mathcal{D}_{UFE}$ and $\mathcal{D}_{UFT}$.

\subsection{Preprocessing Pipeline for Labeled Volumes}
\label{sec:preprocessing}

$\mathcal{D}_{L}$ undergoes the same intensity normalization and resampling pipeline as the unlabeled volumes, with additional steps to align the NIfTI segmentation masks to the corresponding CT volumes. The masks are resampled to a voxel spacing of $(2.0, 1.0, 1.0)$ mm using nearest-neighbor interpolation to preserve discrete labels. Tight 3D bounding boxes are then extracted from the resampled masks via connected-component analysis.


\begin{equation}
\mathcal{B} = (\text{min}_z, \text{min}_y, \text{min}_x, \text{max}_z, \text{max}_y, \text{max}_x)
\end{equation}

\begin{figure}[htbp]
\centering
\includegraphics[width=\columnwidth]{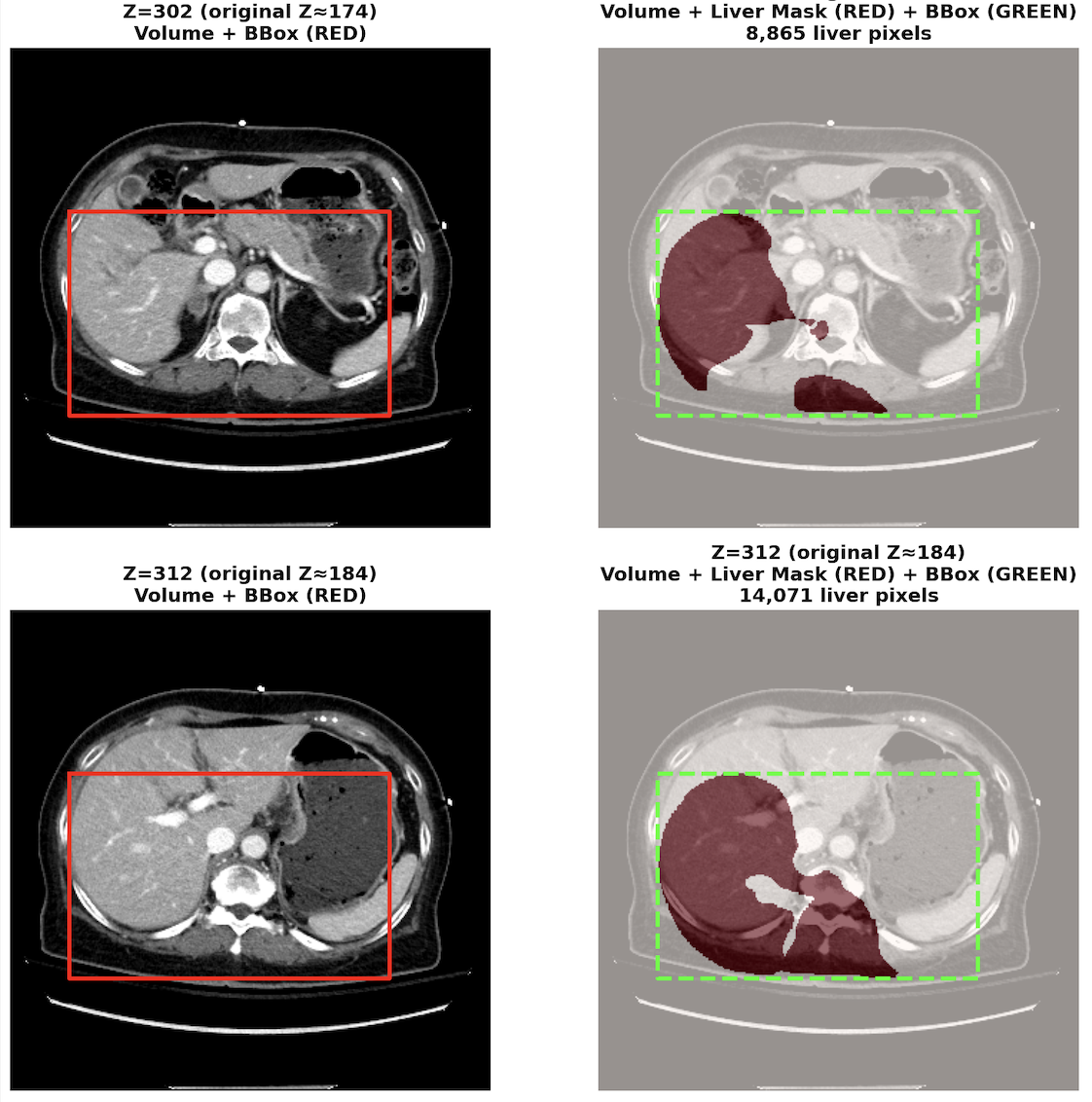}
\caption{Preprocessed CT slices with extracted bounding boxes
(red) and injury masks (red overlay, green boxes).}
\label{fig:preprocessing}
\end{figure}

Figure~\ref{fig:preprocessing} illustrates a labeled sample at different stages of the preprocessing pipeline. The CT volume and corresponding mask are cropped to the resulting bounding box in order to reduce irrelevant background and focus the model on the annotated injury region. The cropped volumes are then standardized to $512 \times 336 \times 336$ voxels through cropping and zero-padding. Each sample is stored in \texttt{.npz} format containing the \textit{volume} (\texttt{float32}), \textit{mask} (\texttt{uint8}), and \textit{box} (bounding-box coordinates).


\section{Methodology}

Inspired by V-DETR for transformer-based 3D detection and by masked image modeling (MIM) for self-supervised representation learning, our method combines medical-domain pretraining with a CT-adapted detection framework. We first pretrain a 3D U-Net encoder using self-supervised learning on unlabeled CT volumes. The pretrained encoder is then integrated into a CT-adapted V-DETR detector, which is further optimized using a semi-supervised teacher--student training strategy.

\subsection{Self-Supervised Pretraining of the Backbone}
To learn robust anatomical features from CT volumes without requiring manual annotations, we pretrain a 3D U-Net encoder using patch-based Masked Image Modeling (MIM). Instead of processing full volumes of size $512\times336\times336$, we extract $128\times128\times128$ patches, which reduces computational cost while enabling broader spatial coverage through data augmentation. This strategy allows us to use $\mathcal{D}_{L}$ together with $\mathcal{D}_{UFE}$, resulting in 1,098 (78/20 train/val) volumes for self-supervised pre-training. Keeping in mind that the 25 volumes are never seen by the pretrained encoder, so that they can be used for testing purposes. 

This subset provides sufficient anatomical diversity while keeping pretraining computationally tractable. The 3D U-Net follows a standard encoder-decoder architecture with skip connections: the encoder progressively downsamples features through 3D convolutions and pooling layers, while the decoder reconstructs the input through transposed 3D convolutions. This architecture enables the network to learn hierarchical multi-scale anatomical representations from volumetric CT data.

\subsection{CT-VDETR: CT-Adapted 3D Detection Model}

\begin{figure}[htbp]
\centering
\includegraphics[width=\columnwidth]{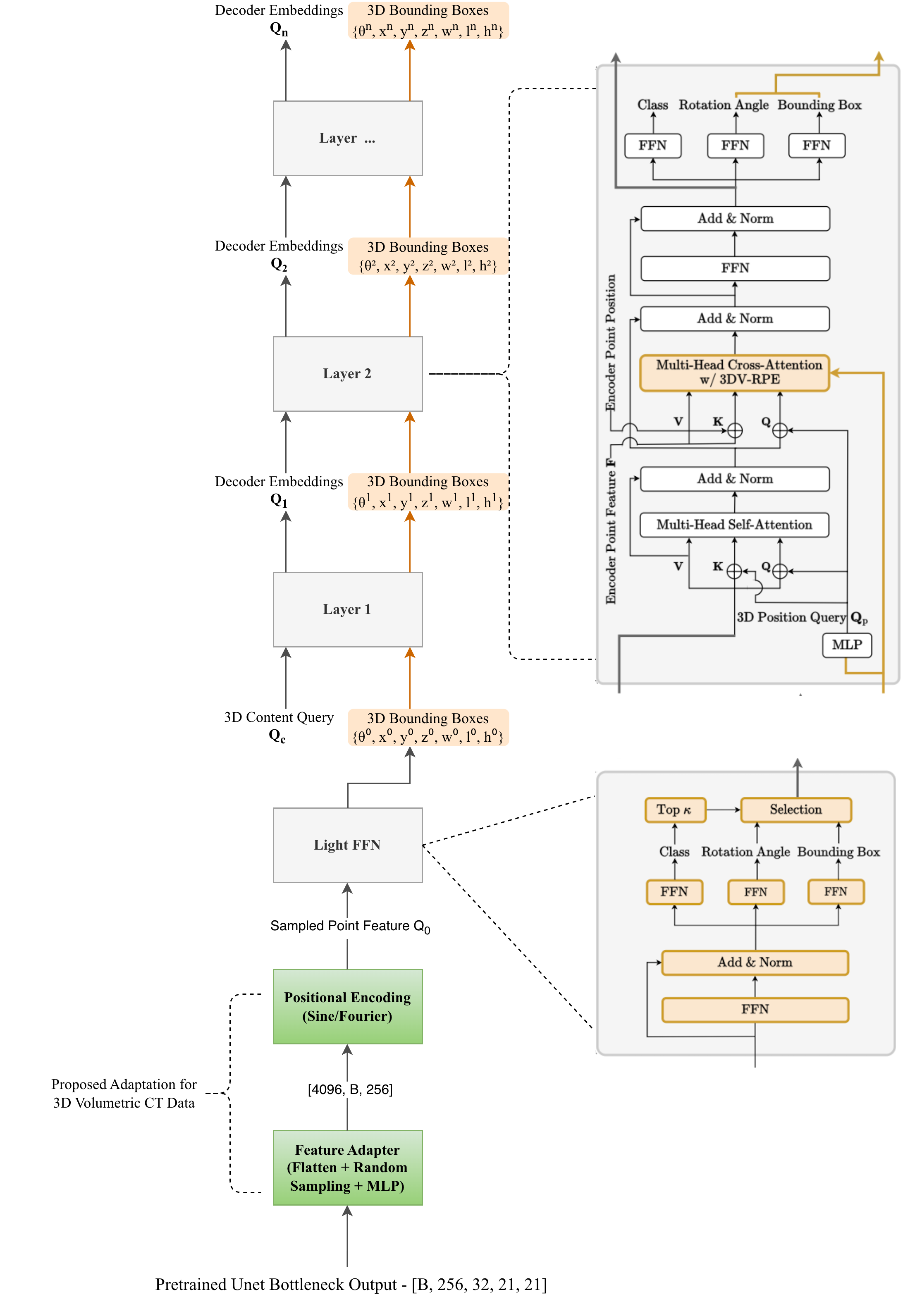}
\caption{Architecture of the proposed CT-VDETR, showing the feature adapter and the 3DV-RPE mechanism, adapted from V-DETR~\cite{shen2023vdetr}.}
\label{fig:vdetr}
\end{figure}

The pretrained 3D U-Net encoder processes $512\times336\times336$ sized voxel CT volumes and generates a $32\times21\times21$ spatial feature grid, with 256-dimensional features at each grid location. To obtain a compact input sequence for the transformer, a feature adapter 
randomly samples 4,096 voxels from the 14,112 spatial locations, reducing memory footprint while maintaining representative spatial coverage. The resulting token sequence is then processed by the V-DETR decoder equipped with 3DV-RPE~\cite{shen2023vdetr}, which provides richer geometric cues than center-based distance measures alone. Figure~\ref{fig:vdetr} illustrates the CT-VDETR architecture and the 3DV-RPE mechanism.

In conventional center-based geometric formulations, spatial relationships are primarily described relative to the predicted box center. For 3D medical imaging, however, such a representation may be insufficient for elongated, deformable, or irregular anatomical structures and injury regions. The 3DV-RPE mechanism addresses this by computing geometric relationships between each spatial token and all eight vertices of the current bounding-box estimate for each object query. These relationships are transformed through multi-layer perceptrons (MLPs) to generate attention biases, which are added to the standard query--key attention scores. As a result, the decoder can incorporate not only semantic similarity but also geometric cues related to inclusion, exclusion, and boundary proximity.

For each query $q$ and voxel position $\mathbf{p}_v$, we compute the offset vectors to all eight vertices of the current query-specific box estimate:

\begin{align}
\Delta\mathbf{P}_i \in \mathbb{R}^{K \times N \times 3},
\end{align}
where $K$ is the number of queries, $N$ is the number of voxel positions (4,096 tokens), and $i \in \{1, \ldots, 8\}$ indexes the box vertices. Each offset is transformed through a non-linear function $F(\cdot)$ and an MLP to produce position bias:

\begin{align}
\mathbf{P}_i = \text{MLP}_i\left(F(\Delta\mathbf{P}_i)\right) \in \mathbb{R}^{K \times N \times h},
\end{align}
where $h$ is the number of attention heads. The total relative position bias is computed by summing across all 8 vertices:

\begin{align}
\mathbf{R} = \sum_{i=1}^{8} \mathbf{P}_i.
\end{align}
This bias augments the standard attention mechanism:

\begin{align}
\mathbf{A} = \text{softmax}\left(\mathbf{Q}\mathbf{K}^T + \mathbf{R}\right),
\end{align}
where $\mathbf{Q}$ represents query embeddings and $\mathbf{K}$ represents key embeddings from the voxel features.

\subsection{Semi-Supervised Training of CT-VDETR} 

To leverage additional unlabeled medical imaging data, we employ semi-supervised learning using teacher--student consistency regularization on $\mathcal{D}_{UFT}$, where the teacher is the same network evaluated without gradients on a weakly augmented view. For each unlabeled volume, weak augmentations (minimal Gaussian noise $\sigma=0.01$, small intensity shifts $\pm 2\%$) generate teacher predictions serving as pseudo-labels without gradient computation. Strong augmentations (larger Gaussian noise $\sigma=0.05$, intensity shifts $\pm 10\%$, Gaussian blur, elastic deformations) are applied to the student input, for which gradients are propagated during training.

The consistency loss comprises three components: $\mathcal{L}_{\text{center}}$ for spatial agreement of box centers ($\mathbf{c}$), $\mathcal{L}_{\text{size}}$ for scale agreement of box sizes ($\mathbf{s}$), and $\mathcal{L}_{\text{cls}}$ for the alignment of class logits ($\mathbf{z}$). The center consistency 
enforces spatial agreement between teacher and student predictions:
\begin{equation}
\mathcal{L}_{\text{center}} = \frac{1}{N}\sum_{i=1}^{N} \|\mathbf{c}_{\mathrm{T}}^{(i)} - \mathbf{c}_{\mathrm{S}}^{(i)}\|_1,
\end{equation} where $\mathbf{c}_{\mathrm{T}}^{(i)}$ and $\mathbf{c}_{\mathrm{S}}^{(i)}$ are the predicted box centers from teacher and student networks for the $i$-th query, 
and $N$ is the number of queries. The size consistency is computed similarly:
\begin{equation}
\mathcal{L}_{\text{size}} = \frac{1}{N}\sum_{i=1}^{N} \|\mathbf{s}_{\mathrm{T}}^{(i)} - \mathbf{s}_{\mathrm{S}}^{(i)}\|_1,
\end{equation} where $\mathbf{s}_{\mathrm{T}}^{(i)}$ and $\mathbf{s}_{\mathrm{S}}^{(i)}$ are the predicted box sizes. Both terms use coordinates normalised by the scene extent $(1024, 336, 336)$ mm. The classification consistency is defined as:
\begin{equation}
\mathcal{L}_{\text{cls}} = \frac{T^2}{N}\sum_{i=1}^{N} \text{BCE}\left(p_{\mathrm{S}}^{(i)},\, p_{\mathrm{T}}^{(i)}\right),
\end{equation}
where $p_{\mathrm{T}}^{(i)} = \sigma(\mathbf{z}_{\mathrm{T}}^{(i)}/T)$, $p_{\mathrm{S}}^{(i)} = \sigma(\mathbf{z}_{\mathrm{S}}^{(i)}/T)$, $\sigma$ is the sigmoid function, and $T=2.0$. Since the detector predicts a single class, each query yields a scalar logit, so a binary cross-entropy replaces a categorical divergence. The total training loss combines supervised detection loss on labeled data with weighted consistency loss on unlabeled data:
\begin{equation}
\mathcal{L}_{\text{total}} = \mathcal{L}_{\text{supervised}} + \lambda(t) \times (\mathcal{L}_{\text{center}} + \mathcal{L}_{\text{size}} + \mathcal{L}_{\text{cls}}),
\end{equation}
where $\lambda(t)$ increases linearly from 0 to 1.0 over the first 10 epochs and remains fixed thereafter. This consistency constraint encourages stable predictions under intensity and geometric perturbations.

\section{Experimental and Evaluation Results}

\subsection{Experimental Setup}

All experiments are conducted on NVIDIA A100 GPUs using the preprocessing protocol described in Section~\ref{sec:preprocessing}. Performance is evaluated using mAP at IoU thresholds of 0.10, 0.25, 0.50, and 0.75, with mAP@0.50 as the primary metric.

The 3D U-Net encoder is pretrained with patch-based MIM on $\mathcal{D}_{UFE}$ and $\mathcal{D}_{L}$ (78/20 train/val) for 50 epochs using Adam ($1\times10^{-4}$). Training uses random $128\times128\times128$ patches, partitioned into $8\times8\times8$ sub-patches with 75\% masking, and optimized with MSE reconstruction loss. The pretrained encoder is then frozen for downstream tasks and the decoder is discarded.

The CT-VDETR detector is trained on $\mathcal{D}_{L}$ (78/20 train/val) using 3D bounding-box annotations and validated on the corresponding validation split. For semi-supervised training, we additionally use samples from $\mathcal{D}_{UFT}$, which are held out during encoder pretraining and preprocessed using the same protocol as the other unlabeled volumes. This disjoint partitioning exposes the detector to anatomical variations not seen during backbone pretraining. 

We train CT-VDETR for 100 epochs using AdamW with a two-phase schedule. In Phase I (epochs 0--20), the pretrained U-Net encoder remains frozen and only the V-DETR decoder and prediction heads are optimized on labeled data with a learning rate of $1\times10^{-4}$. This stabilizes training by allowing the detector to learn the localization task before encoder fine-tuning.
In Phase II (epochs 20--100), the full network is fine-tuned end-to-end. The encoder is unfrozen with a 3-epoch warmup, and its learning rate is increased gradually to $1\times10^{-5}$, while the decoder continues at $1\times10^{-4}$. This stage adapts the pretrained anatomical features to the downstream detection task while reducing catastrophic forgetting.

Consistency regularization is active from the start of training, with $\lambda$ increased linearly from 0 to 1.0 over the first 10 epochs and fixed thereafter. Training uses batch size 1 for labeled data and batch size 2 for unlabeled data.

\subsection{Results}

Table~\ref{tab:detection_test} presents the final test-set evaluation on 25 held-out labeled volumes from $\mathcal{D}_{L}$. The SSL-enhanced model achieves 31.33\% mAP@0.50, compared with 20.45\% for the supervised-only baseline, corresponding to a 1.53$\times$ improvement. The gains are consistent across IoU thresholds up to 0.50, suggesting that self-supervised pretraining and semi-supervised learning substantially improve coarse-to-moderate localization performance under limited labeled data. At the stricter IoU threshold of 0.75, the SSL model reaches 30.95\% barely dropping from 31.33\% yet dramatically outperforms the supervised-only baseline's mere 10.45\% (2.96$\times$ gain). This reveals that semi-supervised learning not only sustains but substantially strengthens precise boundary localization, whereas the baseline catastrophically collapses under strict constraints. The identical mAP values at thresholds 0.10 through 0.50 indicate high geometric stability; consistency losses ($\mathcal{L}_{\text{center}}$ and $\mathcal{L}_{\text{size}}$) constrain the network to produce well-aligned boxes.

\begin{table}[!h]
\centering
\caption{Detection performance comparison on test set.}
\label{tab:detection_test}
\begin{tabular}{lccc}
\toprule
\textbf{Metric} & \textbf{VDETR (no SSL)} & \textbf{VDETR + SSL} & \textbf{Gain} \\
\midrule
mAP@0.10 & 20.45\% & 31.33\% & \textbf{1.53×} \\
mAP@0.25 & 20.45\% & 31.33\% & \textbf{1.53×} \\
mAP@0.50 & 20.45\% & 31.33\% & \textbf{1.53×} \\
mAP@0.75 & 10.45\% & 30.95\% & \textbf{2.96×} \\
\bottomrule
\end{tabular}
\end{table}

\subsection{Ablation Study}

\begin{figure}[H]
\centering
\includegraphics[width=\columnwidth]{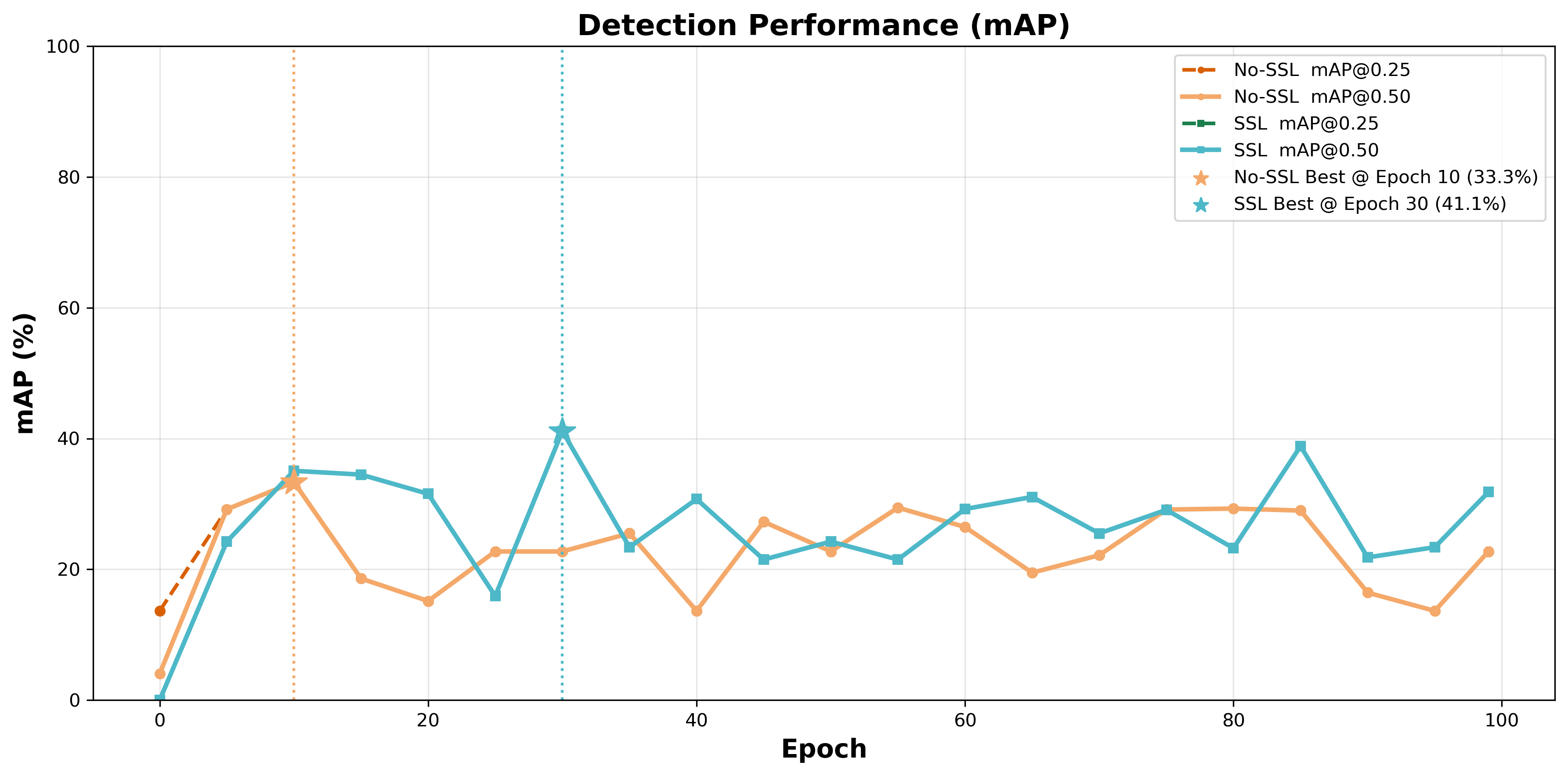}
\caption{Training progression comparison: supervised-only (No-SSL) vs. semi-supervised learning (SSL) with 2,000 unlabeled volumes showing catastrophic collapse vs. stable convergence.}
\label{fig:map_comparison}
\end{figure}

Figure~\ref{fig:map_comparison} compares the training dynamics of the supervised-only baseline (No-SSL) and the proposed semi-supervised variant over 100 epochs. The supervised-only baseline reaches a peak validation performance of 33.3\% mAP@0.50 at epoch 10, but subsequently exhibits significant volatility, dropping to approximately 23\% by epoch 100. In contrast, the semi-supervised model, trained with an additional 2,000 unlabeled volumes and consistency regularization, achieves a peak of 41.1\% mAP@0.50 at epoch 30 and maintains more stable convergence throughout training, ending at approximately 32\% by epoch 100. This corresponds to a 1.23$\times$ improvement over the best supervised-only result and demonstrates substantially better final-epoch performance. Overall, the comparison reveals that incorporating unlabeled data not only improves peak detection performance but also enhances training stability and generalization in the low-label regime, preventing the severe overfitting observed in the supervised-only approach.

\section{Conclusion}

This work presents a label-efficient framework for 3D abdominal trauma detection in CT that combines self-supervised pretraining with semi-supervised learning to address the scarcity of voxel-level annotations. To the best of our knowledge, it is also the first application of a 3D DETR-style detection framework to the RSNA abdominal trauma detection task.

By pretraining a 3D U-Net encoder with masked image modeling and subsequently fine-tuning a CT-adapted V-DETR detector with teacher--student consistency regularization, the proposed approach achieves 41.1\% validation mAP@0.50 and 31.33\% test mAP@0.50 with only 78 labeled training volumes. At the stricter IoU threshold of 0.75, the method achieves a 2.96$\times$ improvement over the supervised baseline, demonstrating that semi-supervised learning particularly strengthens precise boundary localization, which is a critical challenge in trauma detection.

These results validate that combining medical-domain pretraining with semi-supervised optimization is a practical strategy for 3D detection under label scarcity. The dramatic performance gains at IoU threshold of 0.75 suggest that consistency regularization effectively constrains box predictions to well-aligned geometries, overcoming the optimization difficulties that cause supervised-only training to collapse.

Future work will investigate architectural refinements, alternative consistency regularization schemes, evaluation on out-of-distribution datasets, and broader clinical validation, including radiologist-informed assessment and extension to additional abdominal findings or multi-organ trauma detection.

\section*{Acknowledgment}
We sincerely acknowledge the Erlangen National High Performance Computing Center (NHR@FAU) of the Friedrich-Alexander-Universität Erlangen-Nürnberg (FAU) for providing the necessary computing resources that enabled this research.

\bibliographystyle{IEEEtran}
\bibliography{references}

\end{document}